\documentclass[lettersize,journal]{IEEEtran}
\usepackage{amsmath,amsfonts}
\usepackage{algorithmic}
\usepackage{algorithm}
\usepackage{array}
\usepackage[caption=false,font=normalsize,labelfont=sf,textfont=sf]{subfig}
\usepackage{textcomp}
\usepackage{stfloats}
\usepackage{url}
\usepackage{verbatim}
\usepackage{graphicx}
\usepackage{cite}
\usepackage{multirow} 
\usepackage{pifont} 
\usepackage{colortbl} 
\usepackage{xcolor}
\usepackage{threeparttable} 

\usepackage{hyperref}
\hypersetup{
	colorlinks=true,
	linkcolor=cyan,
	filecolor=blue,      
	urlcolor=black,
	citecolor=green,
}

\title{MonoGlass3D: Monocular 3D Glass Detection with Plane Regression and Adaptive Feature Fusion}

\author{Kai Zhang, Guoyang Zhao, Jianxing Shi, Bonan Liu, Weiqing Qi, and Jun Ma, \textit{Senior Member, IEEE}
\thanks{Kai Zhang, Guoyang Zhao, Jianxing Shi, and Weiqing Qi are with the Robotics and Autonomous Systems Thrust, The Hong Kong University of Science and Technology (Guangzhou), Guangzhou 511453, China
(e-mail: kzhang740@connect.hkust-gz.edu.cn; gzhao492@connect.hkust-gz.edu.cn; jshi627@connect.hkust-gz.edu.cn; wqiad@connect.hkust-gz.edu.cn).}
\thanks{Bonan Liu is with the Computational Media and Arts Thrust, The Hong Kong University of Science and Technology (Guangzhou), Guangzhou 511453, China
(e-mail: bliu404@connect.hkust-gz.edu.cn).}
\thanks{Jun Ma is with the Robotics and Autonomous Systems Thrust, The Hong Kong University of Science and Technology (Guangzhou), Guangzhou 511453, China, and also with the Division of Emerging Interdisciplinary Areas, The Hong Kong University of Science and Technology, Hong Kong SAR, China (e-mail: jun.ma@ust.hk).} 
}

\begin{document}

\maketitle

\begin{abstract}

Detecting and localizing glass in 3D environments poses significant challenges for visual perception systems, as the optical properties of glass often hinder conventional sensors from accurately distinguishing glass surfaces. 
The lack of real-world datasets focused on glass objects further impedes progress in this field. To address this issue, we introduce a new dataset featuring a wide range of glass configurations with precise 3D annotations, collected from distinct real-world scenarios.
On the basis of this dataset, we propose MonoGlass3D, a novel approach tailored for monocular 3D glass detection across diverse environments.
To overcome the challenges posed by the ambiguous appearance and context diversity of glass, we propose an adaptive feature fusion module that empowers the network to effectively capture contextual information in varying conditions.
Additionally, to exploit the distinct planar geometry of glass surfaces, we present a plane regression pipeline, which enables seamless integration of geometric properties within our framework. 
Extensive experiments demonstrate that our method outperforms state-of-the-art approaches in both glass segmentation and monocular glass depth estimation.
Our results highlight the advantages of combining geometric and contextual cues for transparent surface understanding. 
The dataset and code will be released at https://github.com/Kai0139/MonoGlass3D.

\end{abstract}


\section{INTRODUCTION}\label{sec:introduction}

The detection and localization of glass surfaces represent a significant and enduring challenge in the field of visual perception \cite{robotic_perception_transp_rev}, primarily attributable to the material's non-distinctive visual appearance and unique transmissive optical properties \cite{gdnet}. These inherent characteristics create a critical gap in environmental awareness for robotic systems. Given the ubiquitous presence of glass structures in modern civil environments, ranging from windows and doors to glass walls and facades, the ability to accurately perceive and localize glass surfaces has become increasingly crucial for the safe and reliable operation of autonomous platforms \cite{liu2024omnicolor}. The inability to detect glass surfaces not only poses substantial safety risks, but also severely limits the operational capabilities of robots in both structured and unstructured environments, making this challenge a pressing issue in the advancement of robotics and autonomous systems.

In some of recent literatures, glass detection is approached as an image segmentation problem \cite{gdnet, eblnet, rfenet},  they are capable of recognizing the presence of glass in images, but lacks crucial 3D information. Practical robotic applications such as path planning \cite{tmechplan} and object manipulation \cite{tmechtransgrasp, tmechmanip} demand a comprehensive 3D understanding of the environment. The primary challenge stems from glass’s inherent transparency \cite{xu2011backimgglass, transdepth}. Robots predominantly rely on cameras and LiDAR sensors, both of which allow visible light and laser beams to pass through glass, these surfaces often appear indistinguishable from the background or even `invisible’ to perception systems. As a result, detecting and localizing glass in 3D space is particularly difficult. Unlike 2D segmentation, which only requires identifying the shape of glass in images, 3D glass detection must also infer spatial position and depth, making the problem substantially more challenging.

\begin{figure}[t]
    \centering
    \includegraphics[width=1\linewidth]{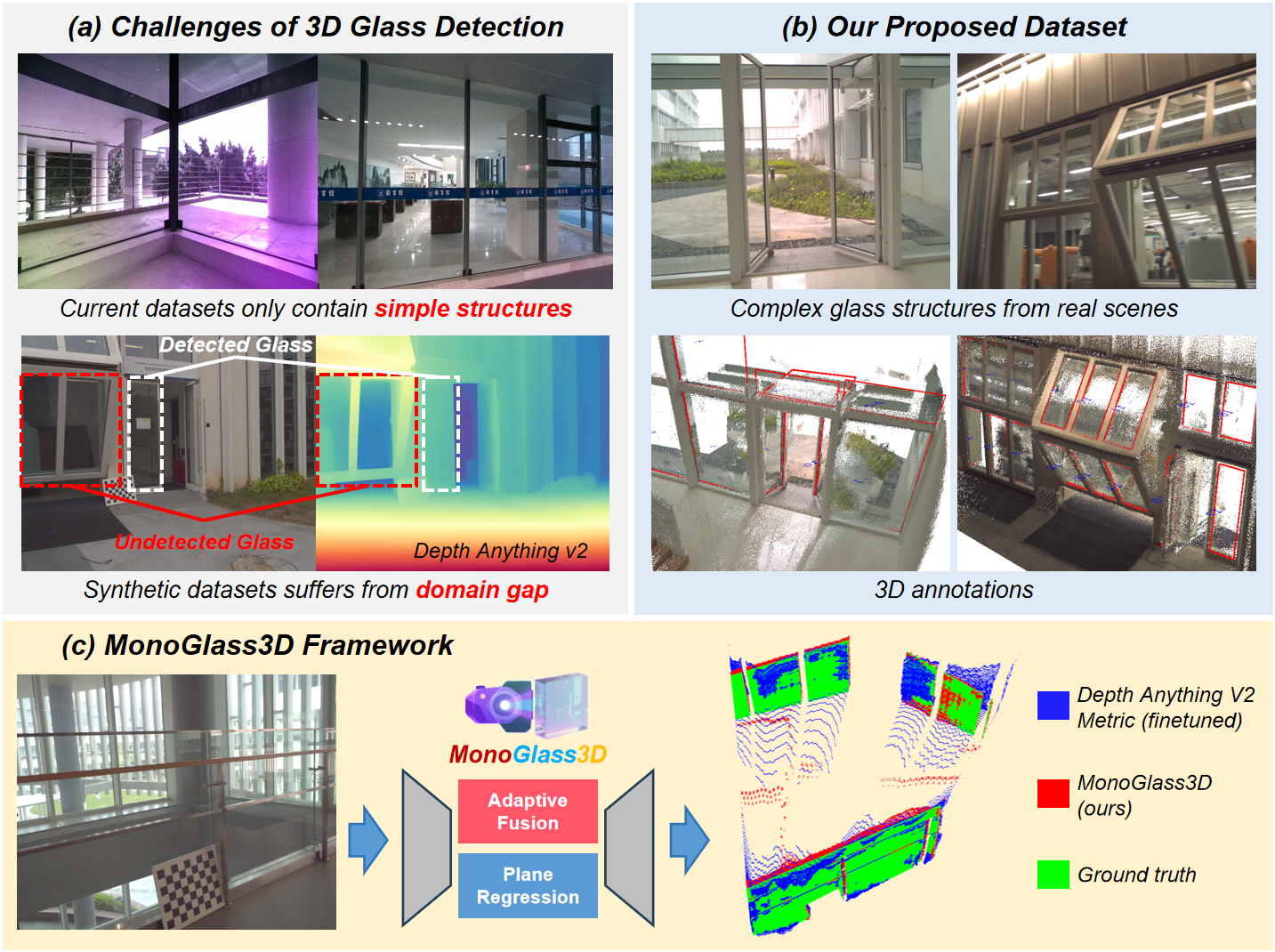}
    \caption{\textbf{Challenges and our proposed method.} (a) Existing 3D glass dataset \cite{gwdepth} only contains simple glass structures with one or two planes. Synthetic datasets suffers from domain gap problems, which makes the models hard to generalize in real-world scenarios, as depicted by depth map from Depth Anything V2 Metric \cite{depth_anything_v2} (pretrained on Hypersim \cite{hypersim}). (b) Our dataset is collected from complex glass structures in real-world. (c) Our MonoGlass3D network achieve superior performance in monocular 3D glass detection when compared to the baseline.}
    \label{fig:intro}
    \vspace{-8pt}
\end{figure}

Some research efforts consider 3D glass detection as an image based depth estimation problem. GWDepth \cite{gwdepth} collects RGB-D data for training a network to estimate the depth of glass walls in civil settings. More recently, Depth Anything V2 \cite{depth_anything_v2} introduces a vision transformer based model that is capable of estimating the depth of glass surfaces. However, the aforementioned approaches have several limitations. First, in 2D glass segmentation, optical features such as blurriness \cite{vbnet2024}, reflections \cite{gwdepth}, and boundaries \cite{eblnet, rfenet} are often utilized to extract contextual information; however, optical features often rely on lighting and may not always be present, while boundary contexts appear as thin lines and are inherently difficult for neural networks to learn. Second, the prior knowledge that glass surfaces are generally planar is not leveraged, overlooking a key geometric constraint that can regulate 3D glass detection. Third, existing 3D glass datasets have notable shortcomings: GWDepth \cite{gwdepth} only includes scenes with one or two glass planes (which are either coplanar or perpendicular, as illustrated in Fig.~\ref{fig:intro}(a)), and Depth Anything V2 is trained on synthetic data \cite{hypersim} only, which introduces domain gap issues and hinders generalization to real-world scenes (Fig.~\ref{fig:intro}(a)).

While our work shares a similar objective with previous studies \cite{gwdepth, depth_anything_v2}, we take a fundamentally different approach. We introduce a new real-world glass dataset and MonoGlass3D, a novel 3D glass detection network. To overcome the limitations of existing datasets and to facilitate efficient collection and annotation of 3D glass surfaces, we build a new real-world glass dataset (Fig.~\ref{fig:intro}(b)) along with a novel data acquisition pipeline. Motivated by the need for more diverse and complex glass configurations, our pipeline leverages a LiDAR-Visual mapping system to produce 3D point clouds containing various glass structures. Glass surfaces are jointly labeled in both 2D images and 3D point clouds. For glass detection, rather than designating specific contextual features, we introduce an adaptive feature fusion module inspired by the centerness concept from \cite{polarmask}, to enhance the network’s adaptability to the diverse shapes present in regions surrounding glass surfaces. Moreover, instead of estimating depth values, our method focuses on regressing plane parameters for glass regions. By reformulating depth estimation as plane parameter regression, we are able to leverage the inherent geometric properties of planes to regulate 3D estimations, resulting in significantly improved 3D glass detection performance, as shown in Fig.~\ref{fig:intro}(c). Extensive experiments on both glass segmentation and glass depth estimation tasks show that our approach achieves superior performance compared to state-of-the-art methods.

In summary, the contribution of this paper is fourfold:
\begin{itemize}
\item  We present a new 3D glass dataset  capturing diverse glass structures and lighting conditions in real-world scenarios, annotated with segmentation masks, 3D plane parameters, and depth maps.
\item We propose an adaptive feature fusion module that leverage centerness maps to enhance the network's adaptability to varying glass contexts.
\item We develop a plane regression pipeline that formulates 3D glass detection into a 3D plane parameter prediction task, enabling the network to benefit from underlying geometric properties.
\item Our method demonstrates noteworthy performance in both glass segmentation and depth estimation, outperforming existing methods with a lightweight framework.
\end{itemize}

\begin{figure*}
    \centering
    \includegraphics[width=0.9\linewidth]{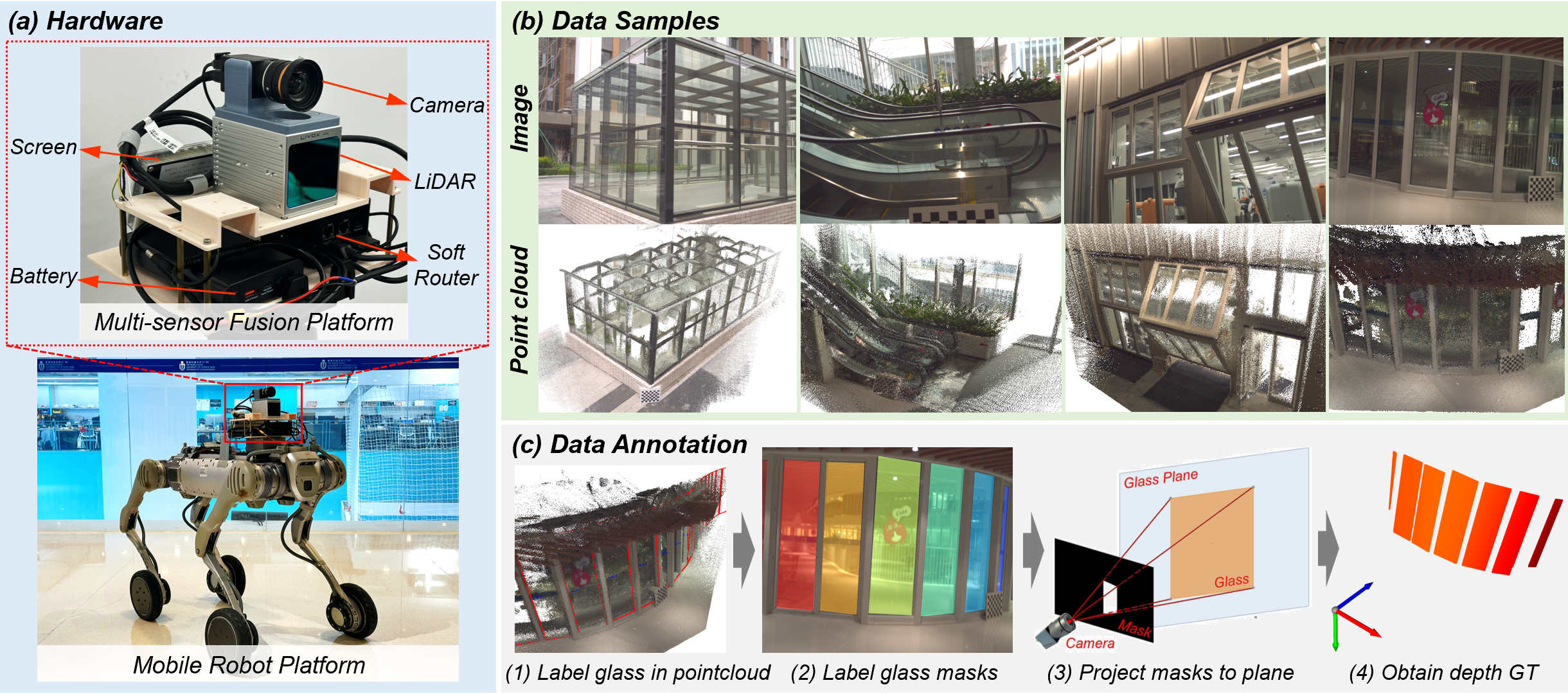}
    \caption{\textbf{Proposed dataset construction procedures.} (a) Data collection device, we integrated a LiDAR and a camera with a wheel-legged robot. (b) Collected scenes with complex glass structures. (c) 3D glass data annotation procedures.}
    \label{fig:dataset_collect_ann}
    \vspace{-10pt}
\end{figure*}

\section{Related Works}

\subsection{2D Glass Detection}
Image-based glass detection raises unique challenges compared to general object detection \cite{sam2} due to the transparent nature of glass elements. Mei et al. \cite{gdnet} pioneers this field by establishing the first benchmark for the glass segmentation problem through introducing GDD, the first glass detection dataset, and GDNet, a glass detection network that identifies glass surfaces by extracting multi-scale contextual features. Although glass surfaces exhibit non-static and ambiguous appearances, certain distinctive features can still be leveraged to detect their presence. For instance, some works \cite{eblnet, rfenet} explicitly use glass boundaries to supervise networks in learning boundary context, while others \cite{vbnet2024, gsd, liu2024multi_vgsd} extract glass reflections to enhance the network’s capability of recognizing glass. In addition, semantic labels have been utilized to guide networks in learning glass context features associated with semantic segmentation \cite{gsds, depth_anything_v2}. To better distinguish glass from the background, different sensor modalities have been incorporated in previous works. Polarization imaging sensors, which capture the linear polarization states of pixels, have been employed in \cite{deeppolartrans, Mei_2022_CVPR_pgsnet}, and thermal imaging sensors have also been explored \cite{rgbtglass}. However, these alternative sensors are not as accessible as RGB cameras, which hinders their application in realistic scenarios.

\subsection{3D Glass Detection}
Unlike glass segmentation in images, it is much harder to accurately measure the position of glass in 3D space. Common 3D sensors, such as depth cameras and LiDARs, both rely on measuring the time of flight of transmitted infrared (IR) signals. However, similar to visible light, IR signals often penetrate the glass surface and reflected by the object behind glass, the reflected signal is also unreliable because of light refraction. To avoid these difficulties, synthetic data becomes a practical and efficient approach to generate annotated 3D data for glass. Synthetic dataset is employed in \cite{transdepth, cleargrasp} to train depth estimation network for small transparent objects, Depth Anything V2 \cite{depth_anything_v2} proposes a synthetic-to-real knowledge transfer pipeline with a teacher model pre-trained on Hypersim \cite{hypersim}, a large scale synthetic dataset featuring hundreds of indoors scenes with glass surfaces.

The shape distribution of glass surface in real world is highly long tailed, with the majority being flat surfaces. For flat glass, dense depth maps can be interpolated from a few points. GWDepth \cite{gwdepth} uses a depth camera to capture the depth values at glass boundaries and and interpolate pixel-wise glass depths. However, this requires manual annotation of reliable pixels in each frame. Additionally, while the interpolability of flat glass surfaces is a valuable geometric constraint for glass detection, their approach does not fully leverage this property.
\vspace{-10pt}

\subsection{3D Perception with Planar Constraints}

Among geometric primitives, planes are especially useful as they regulate unconstrained 3D points to lie on a 2D surface. Recent advances in deep learning have enabled significant progress in 3D perception leveraging planar constraints. PlaneNet \cite{planenet_cvpr2018} introduces the first end-to-end neural network for piecewise planar reconstruction from a single RGB image, they estimate plane parameters and segmentation masks to yield structured depth maps with high accuracy. Similarly, PlaneRecover \cite{planerecover_eccv2018} proposes a convolutional neural network framework that exploits a novel plane structure-induced loss, allowing the network to simultaneously predict planar segmentation without requiring explicit plane annotations. Building upon these ideas, PlaneRCNN \cite{planercnn_cvpr2019} adapts a Mask R-CNN architecture for plane detection and jointly refines segmentation masks by enforcing multi-view consistency during training. Moving beyond explicit plane detection, P3Depth \cite{P3Depth_cvpr2022} introduces a piecewise planarity prior to monocular depth estimation, learning to aggregate information from coplanar pixel groups through a two-headed network structure that adaptively fuses predictions, resulting in depth maps with sharp geometric boundaries and leading performance on standard benchmarks. These advances collectively demonstrate the effectiveness of integrating planar priors into deep networks for improved 3D perception from single RGB images. Most of the related works treat plane parameter estimation as an auxiliary task to improve object detection or depth estimation. In contrast, our approach is designed to predict the plane parameters of glass surfaces, which tackles 3D glass detection problem from a new perspective.

\section{Monocular 3D Glass Detection Dataset}

\subsection{Hardware Setup}

Compared to depth cameras \cite{transdepth, gwdepth}, LiDAR systems generally provide more accurate distance measurements. However, conventional LiDAR produces inherently sparse data and does not deliver depth values for every pixel within the Field of View (FoV), limiting its utility for dense 3D reconstruction of transparent surfaces such as glass. To address this, we select the Livox-Avia, a compact, non-repetitive scanning LiDAR. Unlike repetitive scanning LiDARs, which produce a fixed point cloud pattern, non-repetitive scanning allows point cloud density to accumulate over time as the sensor moves slowly during data acquisition, resulting in more comprehensive 3D coverage of glass surfaces. We configure the LiDAR to operate in single-first return mode to ensure stable and reliable mapping, as multi-return modes are usually incompatible with LiDAR-Visual SLAM systems.

For the RGB camera, we select the MV-CU013-80GC global shutter camera from HIKROBOT, which captures images at resolution of $1280\times1024$. A $6\,\mathrm{mm}$ lens is deliberately chosen to ensure the camera’s FoV is slightly smaller than that of the LiDAR. This design guarantees that all glass surfaces captured in the images can be reliably mapped to the corresponding 3D LiDAR point clouds, thereby ensuring consistent cross-modal annotation.

To optimize image quality under varying lighting conditions, we tailor the camera settings to different environments. In indoor environments, automatic exposure is enabled to maintain uniform brightness across frames. In outdoor environments, where lighting can fluctuate rapidly, we disable automatic exposure and instead dynamically adjust the exposure time in real-time according to brightness changes of adjacent frames, effectively preventing overexposure and ensuring high-quality image capture regardless of ambient brightness.

For sensor integration, we adopt the configuration in FAST-LIVO2 \cite{fastlivo2}, as illustrated in Fig.~\ref{fig:dataset_collect_ann}(a). Unlike FAST-LIVO2 that synchronizes sensors using external PPS signals, we use a soft router with multiple Ethernet interfaces that support the IEEE1588 Precision Time Protocol (PTP). This configuration enables precise timestamp synchronization between image frames and LiDAR scans through a PTP master device, without requiring any external signal sources. Extrinsic calibration between the LiDAR and camera is performed using a chessboard target, wherein the transformation matrix is computed by matching chessboard corners detected from LiDAR intensity images and RGB images, ensuring accurate spatial correspondence between modalities.

To further enrich the diversity and realism of our dataset, we mount the synchronized sensor suite on the wheel-legged Unitree B2-W robot, which offers high mobility and stability over challenging terrains, such as stairs and curbs. This hardware integration enables data collection in both handheld and robot-carried modes, allowing for comprehensive coverage of complex environments and facilitating the construction of a realistic and versatile 3D glass dataset.

\subsection{Data Collection}

In this work, we use FAST-LIVO2 \cite{fastlivo2} to produce 3D scans of glass and surrounding environments. FAST-LIVO2 is a state-of-the-art open sourced multi-modal SLAM system, which employs sensor fusion techniques to combine LiDAR, RGB camera, and IMU data to estimate the camera poses while generating a colored point cloud map. To ensure a wide diversity of glass geometries, we capture data in a variety of locations, including corridors, cafes, libraries, offices, glass doors, glass walls, ceilings, escalators, etc. In total, we scan 50 distinct scenes featuring a broad range of glass configurations, as exemplified in Fig.~\ref{fig:dataset_collect_ann}(b).

\subsection{Data Annotation}

The data annotation mainly consists of 4 steps, as shown in Fig.~\ref{fig:dataset_collect_ann}(c). First, for each scene, after we obtain the point cloud map, we annotate the 3D positions of glass surfaces in the point cloud using thin bounding boxes. The vertices of bounding boxes are then transformed to the camera coordinate frame by applying spatial transformations: 
\begin{equation}
\label{eq:vertex_tf}
P_c = T_{cw}P_w
\end{equation}
where $P_c$ and $P_w$ denote the bounding box vertex coordinates in camera and world frames, respectively, $T_{cw}$ is the transformation matrix mapping world coordinates to camera frame.

Then, we fit the 3D bounding boxes to planes. A plane in 3D can be described by 4 variables characterized by:
\begin{equation}
\label{eq:plane}
    \mathbf{n \cdot v} + d = 0
\end{equation}
where $\mathbf{n}$ is the 3D plane normal vector (contains 3 variables), and $\mathbf{v}$ is arbitrary point on the plane, $d$ is the interception, with magnitude equals to the distance from origin to the plane. Given 3D positions of bounding box vertices $P_c$, we can fit the vertices to a set of plane parameters using least squares approximation:
\begin{equation}
\label{eq:leastsq}
\begin{split}
\mathbf{\hat n} &= (P_c^T P_c)^{-1}P_c^T \begin{bmatrix}
    -1 \end{bmatrix}^{3\times1} \\
\mathbf{n} &= \frac{\mathbf{\hat n}}{\|\mathbf{\hat n}\|}, \quad d = \frac{1}{\|\mathbf{\hat n}\|}
\end{split}
\end{equation}

After glass instances are registered as 3D planes, we annotate glass in images. Segmentation masks for glass surfaces are annotated on selected image frames from each scan. 

Finally, after we obtain 2D masks and their corresponding 3D plane parameters, the depth value for each pixel within the mask is computed by projecting each 2D mask to its associated 3D plane:
\begin{equation}
\label{eq:intersect}
\mathrm{depth} = d\left( \mathbf{n}^T K^{-1} \begin{bmatrix}u_x & u_y & 1\end{bmatrix}^T \right)^{-1}
\end{equation}
where $K$ is the camera intrinsic matrix, $u_x$ and $u_y$ are the pixel coordinate. This projection process is depicted as the last two steps of Fig.~\ref{fig:dataset_collect_ann}(c).

\subsection{Dataset Information}

\begin{figure}[h]
\vspace{-10pt}
    \centering
    \includegraphics[width=1\linewidth]{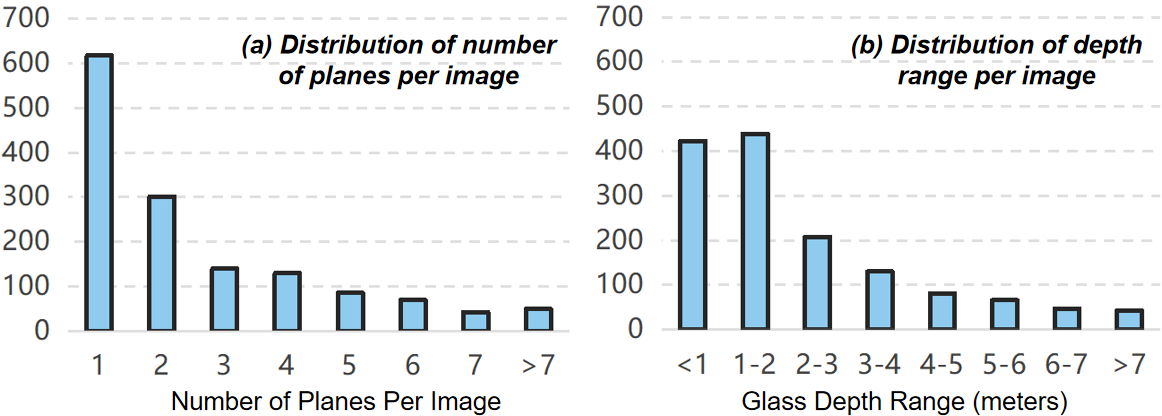}
    \caption{\textbf{Statistics of proposed dataset.} (a) Distribution of number of planes in individual images. (b) Distribution of the range of glass depths in individual images.}
    \label{fig:data_stats}
\vspace{-5pt}
\end{figure}

The proposed labeling pipeline enables efficient generation of precise depth ground truth for a large number of images with minimal manual effort. Building on this capability, our dataset exhibits a substantially greater diversity in glass planar configurations. Eventually, we obtain 1437 annotated frames, with 1,070 images allocated to the training set and 367 images to the validation set. As illustrated in Fig.~\ref{fig:data_stats}(a), the number of glass planes per image in our dataset ranges from 1 up to 10, providing a broader spectrum of geometric complexity. Furthermore, as shown in Fig.~\ref{fig:data_stats}(b), the range of glass depths within individual images spans from $0.07\,\mathrm{m}$ to $17.76\,\mathrm{m}$, increasing the challenge of 3D estimation. Additionally, unlike previous approaches, 21 out of our 50 scenes are collected at night, introducing a wide variety of illumination conditions and further challenging the robustness of glass detection methods.

\begin{figure*}
    \centering
    \includegraphics[width=1\linewidth]{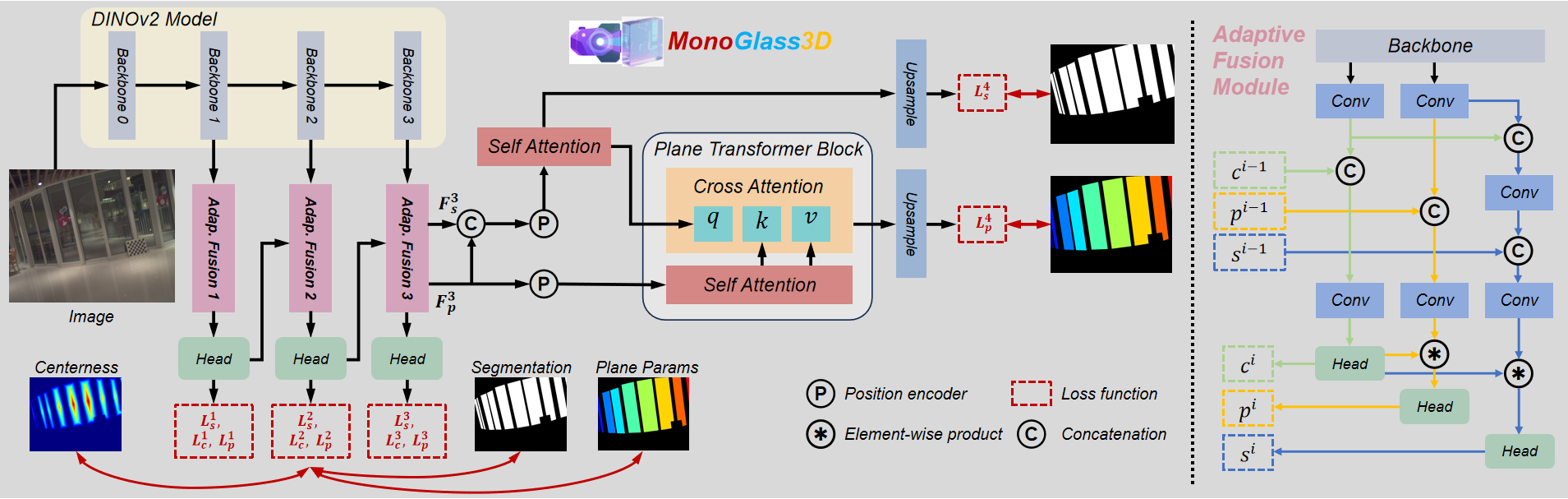}
    \caption{\textbf{Proposed MonoGlass3D framework.} Our network consists DINOv2 backbone, centerness adaptive fusion module, and attention layers. Our network jointly performs glass segmentation and plane regression, which are supervised by segmentation loss $L_s$, centerness loss $L_c$, and plane loss $L_p$.}
    \label{fig:network}
    \vspace{-10pt}
\end{figure*}

\section{Methodology}

\subsection{Framework Overview}

Our proposed MonoGlass3D is designed to perform both glass segmentation and plane regression simultaneously with a simple overall architecture. As shown in Fig.~\ref{fig:network}. Specifically, the outputs from the last three layers of the DINOv2 encoder, each with dimensions $C\times\frac{H}{14}\times\frac{W}{14}$, are processed by the adaptive fusion modules. The adaptive fusion module first reduces the feature channel dimension to 256. Then the prediction results from the previous level, namely centerness $c^{i-1}$, plane parameters $p^{i-1}$, and segmentation $s^{i-1}$, are concatenated with the features and further refined using convolution layers. Centerness $c^i$ is predicted first; it is then fused with the segmentation and plane features by element-wise multiplication, enhancing the features according to the geometric properties captured by centerness. The adapted features, $F_s$ for segmentation and $F_p$ for plane regression, are fed into their respective prediction heads to generate segmentation masks and plane parameter predictions at $\frac{H}{14}\times\frac{W}{14}$ resolution for each encoder layer. Features from the last layer, $F_p^3$ and $F_s^3$ are concatenated, augmented with position encoding, then passed to self-attention layers for further refinement, then the final glass segmentation mask is upsampled via transpose convolution layers. The output plane features from the last adaptive fusion layer, $F_p^3$, is combined with position encoding, and further refined using self-attention layers with transformer encoders, followed by cross-attention with the segmentation features using transformer decoder layers. Finally, the refined features are up sampled to predict the plane parameters via transpose convolutions.

\subsection{Plane Parameter Regression}

\begin{figure}[h]
    \centering
    \includegraphics[width=0.95\linewidth]{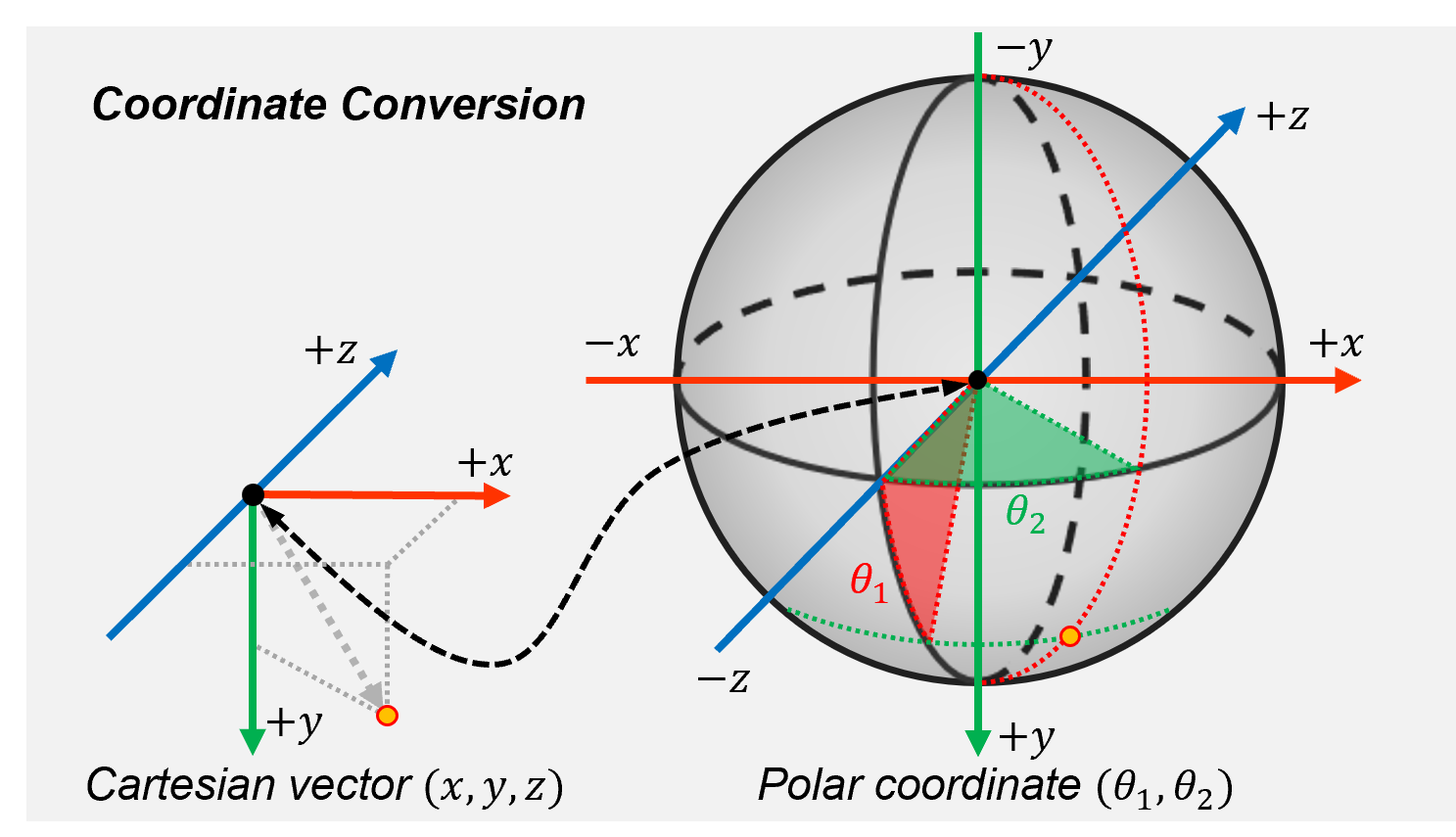}
    \caption{\textbf{Coordinate conversion of plane normal vector.} Cartesian vector $(x,\,y,\,z)$ is mapped to its corresponding polar coordinate $(\theta_1,\space \theta_2)$.}
    \label{fig:plane_ang}
    \vspace{-10pt}
\end{figure}

As mentioned, a 3D plane can be defined by four variables: a 3D normal vector and interception. However, directly regressing four unconstrained variables is not ideal. The 3D space has 6 Degrees of Freedom (DoF), three for translation and three for rotation, but the intrinsic dimensionality of a plane is only 3 DoF. This is because two translation axes and one rotation axis (along the plane direction) are inherently undefined for a plane. Consequently, representing a plane using four free variables leads to over-parameterization. Usually, the plane normal vector is normalized to fix the vector length DoF. To avoid the need for post hoc normalization, we re-map the plane normal to a polar coordinate system with two angles, $[\theta_1,\, \theta_2]$, as shown in Fig.~\ref{fig:plane_ang}. The algebraic conversion is defined as:
\begin{equation}
\begin{aligned}
\theta_1 &=
\begin{cases}
\arccos\left(\frac{r_{xz}}{r}\right), & \text{if } y > 0, \\
-\arccos\left(\frac{r_{xz}}{r}\right), & \text{otherwise}
\end{cases} \\
\theta_2 &=
\begin{cases}
\pi - \arccos\left(\frac{z}{r_{xz}}\right), & \text{if } x > 0, \\
-\arccos\left(\frac{z}{r_{xz}}\right), & \text{otherwise}
\end{cases}
\end{aligned}
\end{equation}
where $r_{xz}$ is the vector magnitude in $x$ and $z$ axes, $r_{xz}=\sqrt{r_x^2+r_z^2}$, and $r=1$ for unit vectors. When the plane is perpendicular to $z$ axis, $\theta_1 = \theta_2 = 0$, which corresponds to normal vector $[0,\space0,\space-1]$.

The plane normal vector can be regulated in the $-z$ hemisphere, this is because according to (\ref{eq:plane}), when the normal vector $\mathbf{n}$ points to the $+z$ direction, we can flip the sign of both $\mathbf{n}$ and $d$ to yield the equivalent plane representation. The polar representation effectively reduces the parameter space, and narrows down the range of plane parameters to $\theta_1,\theta_2 \in [-\frac{\pi}{2}, \space \frac{\pi}{2}]$ and $d \in \mathbb{R}$. Compared with conventional approaches that estimate four parameters (three for the normal vector and one for the intercept), our angular representation reduces plane estimation to three parameters, $[\theta_1 \quad \theta_2 \quad d]$, and eliminates the need for additional normalization. In our network, the prediction heads estimate the angles $\theta_1$ and $\theta_2$ using $\tanh$ activation, up-scaled by $\frac{\pi}{2}$, while the interception $d$ is up-scaled by a factor of $5$.

For planes with a same normal direction, the value and sign of interception $d$ vary with the plane's position within the image, making positional encoding critical for accurate inference. Since the DINOv2 backbone utilizes transformer encoder layers with positional encoding, we directly apply adaptive fusion module to the outputs of the last three encoder layers to predict the plane parameters. Additionally, we enhance both the segmentation features $F_s^3$ and the plane features $F_p^3$ with extra positional encoding. We then refine $F_p^3$ by applying self-attention through transformer encoder layers, providing global context, and followed by cross-attention using transformer decoder layers, where segmentation features serves as the query embedding to localize glass across the image. The final plane parameter estimations are up sampled and projected to $3 \times H \times W$, with 3 channels corresponding to $[\theta_1 \quad \theta_2 \quad d]$.
\vspace{-5pt}
\subsection{Adaptive Feature Fusion with Centerness}

In real-world scenarios, glass typically exhibit regular shapes such as rectangles, trapezoids, and ellipses. Geometric context is therefore a crucial property for both plane estimation and glass segmentation. Previous works have leveraged boundary information \cite{eblnet, rfenet} to enhance contextual learning; however, boundaries are represented as thin lines in images, making them challenging to be learned efficiently. 

In our approach, we adopt the concept of centerness introduced by \cite{fcos_centerness} and \cite{polarmask}. Centerness was originally designed for instance segmentation, it is quantified by the normalized distance from a pixel to the boundary of an object instance. Specifically, for an pixel within an instance, centerness is defined by:
\begin{equation}
\label{eq:centerness}
    \mathcal{C} = \sqrt{\frac{d_{min}}{d_{max}}}
\end{equation}
where $d_{min}$ and $d_{max}$ are the shortest and longest distance to the instance contour from that pixel, respectively.

\begin{figure}
     \centering
     \includegraphics[width=0.9\linewidth]{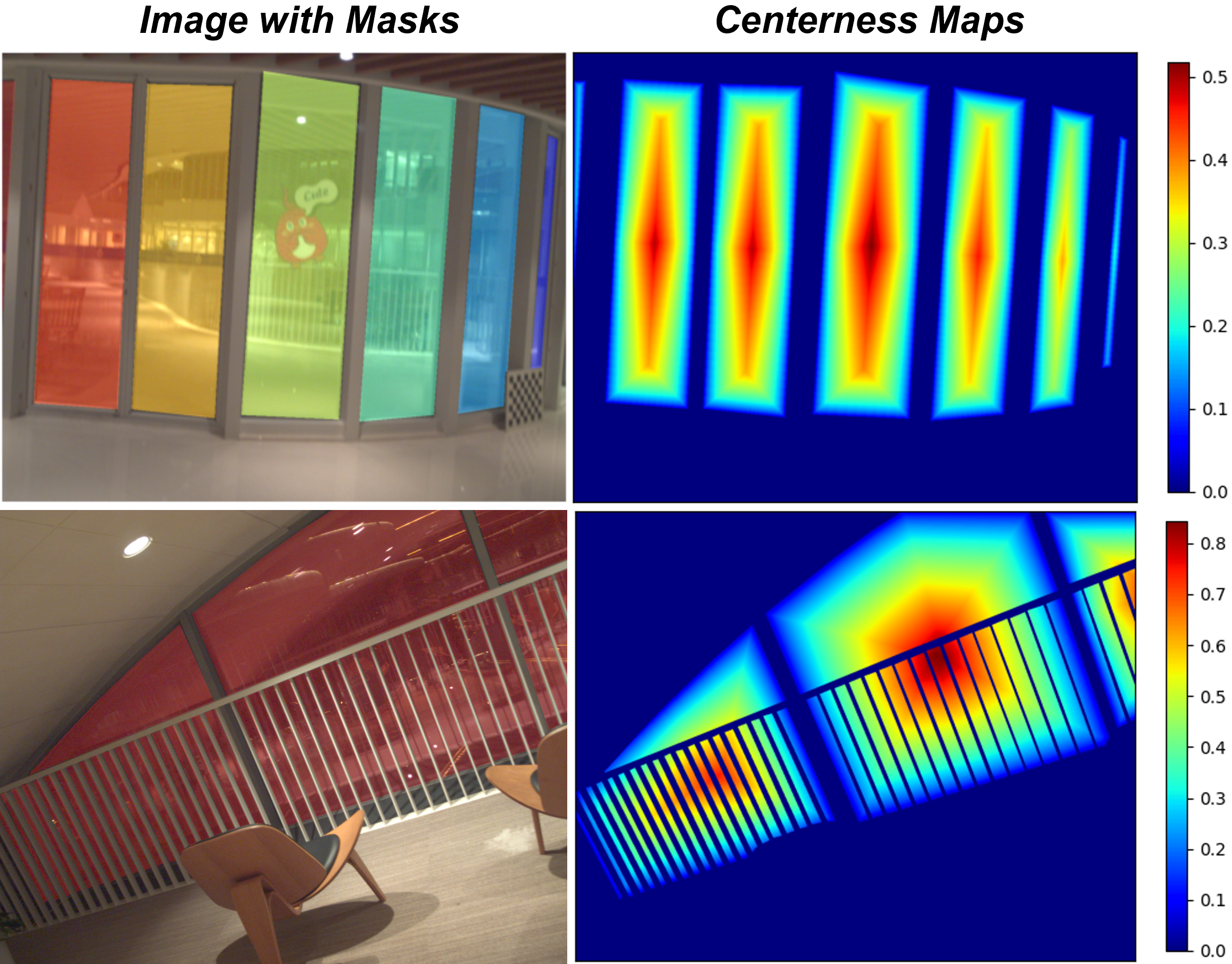}
     \caption{\textbf{Centerness map examples.} Image with masks (left) and corresponding centerness maps (right).}
     \label{fig:centerness}
     \vspace{-10pt}
 \end{figure}

As shown in Fig.~\ref{fig:centerness}, the centerness distribution closely reflects the geometric shape of the mask. Typically, the centerness forms a smooth, pyramid-like distribution, instances with aspect ratios closer to $1$ exhibit higher peak values at their centers. Compared to boundary context, centerness provides a much smoother signal that is easier for neural networks to learn and generalize from, while still capturing rich geometric context at the instance level.

In our network, predicted centerness is utilized to enhance both segmentation and plane features in adaptive fusion modules, but we multiply with $1 - \mathcal{C}$, because the features extracted from the glass surface is less important and possibly distracting, we would like the network to focus more on the context near the boundaries, which are more useful for extracting the geometric properties of glass.
Centerness is integrated by feature fusion through: 
\begin{equation}
\label{eq:fusion}
    F_{fused}=F \ast (1-\mathcal{C)}
\end{equation}
where $\ast$ denotes element-wise multiplication.

By integrating centerness in this manner, our model adaptively emphasizes features according to their geometric context, thereby improving both segmentation accuracy and plane parameter estimation.

\subsection{Loss Function}
\label{subsec:loss}

The training of our network is supervised by three types of ground-truth: centerness, segmentation masks, and plane parameters:
\begin{equation}
\label{eq:loss_overall}
    L = L_c + L_s + l_p
\end{equation}
where $L$ is the total loss, $L_c$ denotes the centerness loss, it is computed using binary cross-entropy (BCE) loss. For segmentation, we employ a composite loss $L_s$, that combines BCE loss and Intersection over Union (IoU) loss:
\begin{equation}
\label{eq:seg_loss}
    L_s = 0.5L_{bce} + L_{iou}
\end{equation}
The plane loss (denoted as $L_p$) is designed to more effectively supervise the learning of plane parameters. Previous works such as \cite{planenet_cvpr2018, P3Depth_cvpr2022} use depth loss to supervise plane parameter estimation. However, this is not appropriate, because depth is the projection of 3D points onto the $z$-axis from the camera’s perspective, it is distributed non-uniformly for different plane orientations. 

\begin{figure*}
    \centering
    \includegraphics[width=1.0\linewidth]{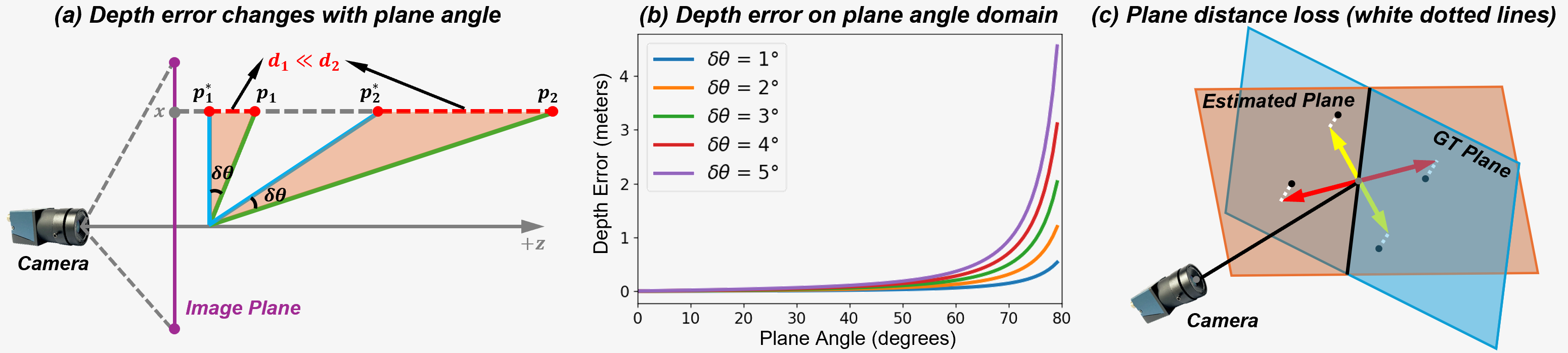}
    \caption{\textbf{Drawbacks of depth loss and proposed plane distance loss.} (a) Depth error at same angular error varies largely depending on plane angles. (b) Plot of depth error on the plane angle domain. (c) Plane distance loss, measures the distance between 3D planes (white dotted lines).}
    \label{fig:plane_loss}
    \vspace{-10pt}
\end{figure*}

To further elaborate this issue, as shown in Fig.~\ref{fig:plane_loss}(a), consider an image pixel at $x$, for a fixed angular error $\delta\theta$ between predicted planes $p_1,\space p_2$, and ground-truth planes $p_1^\ast, \space p_2^\ast$, the depth errors $d_1$ and $d_2$ can vary significantly. As depicted in Fig.~\ref{fig:plane_loss}(b), the depth error increases nonlinearly with the plane angle (when the plane interception fixed at $1$), leading to highly uneven error distributions in the plane angle domain. Therefore, directly employing depth error as the supervision signal for plane parameters can result in suboptimal training.

To address this issue, we propose the plane distance loss. Given a plane estimation at a particular pixel location $(u_x, \space u_y)$, we first determine its corresponding 3D position $p$ by projecting the pixel onto the estimated plane. Next, we select 4 points on this plane,  each at an equal distance from $p$ and oriented along orthogonal directions, as illustrated in Fig.~\ref{fig:plane_loss}(c). The plane distance loss is defined as the sum of the distances from these four points to the ground-truth plane (represented by the white dotted lines). Unlike depth error, these distances represent the true geometric discrepancy between the two planes and are invariant with respect to the camera viewpoint and plane orientation.

In addition to the plane distance loss, we add an L1 loss directly on the plane parameters. Thus, the plane loss for each pixel is the sum of the plane distance loss and the plane parameter loss:
\begin{equation}
    L_p = L_{param} + L_{dist}
    \label{eq:plane_loss}
\end{equation}

Since different glass instances occupy varying areas in the image, we normalize the plane loss at the instance level to mitigate the effect of mask size. For an image with $N$ instances, where the $i$-th instance contains $M_i$ pixels,  the plane loss is averaged across all pixels and all instances, which can be defined as:
\begin{equation}
\label{eq:plane_loss_norm}
    L_p = \frac{1}{N}\sum_{i=0}^N\frac{1}{M_i}{\sum_{j=0}^{M_i} L_{p,(i,j)}}
\end{equation}

Segmentation and plane losses are computed for all output stages, while centerness loss is only evaluated at cascade layers. To balance the contributions from different stages, we apply stage-specific weights to each loss, defined as:
\begin{equation}
\label{eq:loss_weighted}
\begin{split}
    L_p &= 0.1L_p^1 + 0.1L_p^2 + 0.2L_p^3 + 0.6L_p^4 \\
    L_s &= 0.1L_s^1 + 0.1L_s^2 + 0.2L_s^3 + 0.6L_s^4 \\
    L_c &= 0.2L_c^1 + 0.3L_c^2 + 0.5L_c^3
\end{split}
\end{equation}

\section{Experiments}

\subsection{Training Setup}

Our network is implemented with PyTorch \cite{pytorch}. For the backbone, we employ DINOv2 ViT-S, initialized with pretrained weights from Depth Anything v2 (ViT-S). Our network is trained and evaluated at resolution of $504\times630$. We use AdamW optimizer, the backbone learning rate is set to $5\times10^{-6}$, all other layers are set to $5\times 10^{-5}$. We adopt a learning rate scheduler which decays the learning rate by a factor of 0.95 if the loss does not decrease for 8 epochs. Our model is trained for 230 epochs on 4 Nvidia RTX 4090 GPUs.

\subsection{Evaluation Metrics}

We demonstrate the performance of our approach by comparing with recent works on two metrics: glass segmentation and depth estimation.

For segmentation performance, we use the following metrics: IoU, defined as \small$\frac{TP}{TP+FP+FN}$\normalsize, it measures the overlap between prediction and ground-truth. Mean Absolute Error (MAE), defined as $\frac{1}{|\mathcal{I}|} \sum_{i \in \mathcal{I}} |y_i^\ast - y_i|$, where $y_i$ is the predicted probability at pixel $i$ in image $\mathcal{I}$, $y_i^\ast$ is the ground-truth. F1 score, defined as \small$\frac{2(\text{Precision}\, \times \, \text{Recall})}{\text{Precision} \space + \space \text{Recall}}$\normalsize, it is a harmonic mean of Precision and Recall, where Precision is \small$\frac{TP}{TP+FP}$\normalsize and Recall is \small$\frac{TP}{TP + FN}$\normalsize. Balanced Error Rate (BER), defined as \small$1 - 0.5(\frac{TP}{TP+FP} + \frac{TN}{TN+FN})$\normalsize, it is the average of errors on both True and False predictions.

Although our network outputs plane parameters, we report results using depth estimation metrics for more intuitive comparison with previous works. We use the following depth evaluation metrics: Mean absolute error (MAE), similar to segmentation, except here $y$ denotes depth instead of segmentation probability. Root Mean Squared Error (RMSE), defined as 
\small$\sqrt{\frac{1}{|\mathcal{I}|}\sum_{i \in \mathcal{I}}(y_i^\ast-y_i)^2}$\normalsize, which is a quadratic mean of the errors. Absolute Relative Error (ARE), defined as $\frac{1}{|\mathcal{I}|} \sum_{i \in \mathcal{I}}{|y_i^\ast - y_i|}/{y_i^\ast}$, it measures the error normalized by the true value. Accuracy with Threshold, defined as $S_t = {\{ i | \max{(\frac{y}{y^\ast}, \space \frac{y^\ast}{y}) < t} \}}/{|\mathcal{I}|}\}$, It measures the proportion of pixels where the ratio of estimated to ground-truth depth is below threshold $t$. Following previous works \cite{depth_anything_v2, gwdepth}, we use thresholds of $1.25$, $1.25^2$, and $1.25^3$. These metrics are denoted as $\sigma_1 = S_{1.25}$, $\sigma_2 = S_{1.25^2}$, and $\sigma_3 = S_{1.25^3}$, respectively.

\subsection{Performance Evaluation}

\begin{figure}
    \centering
    \includegraphics[width=1\linewidth]{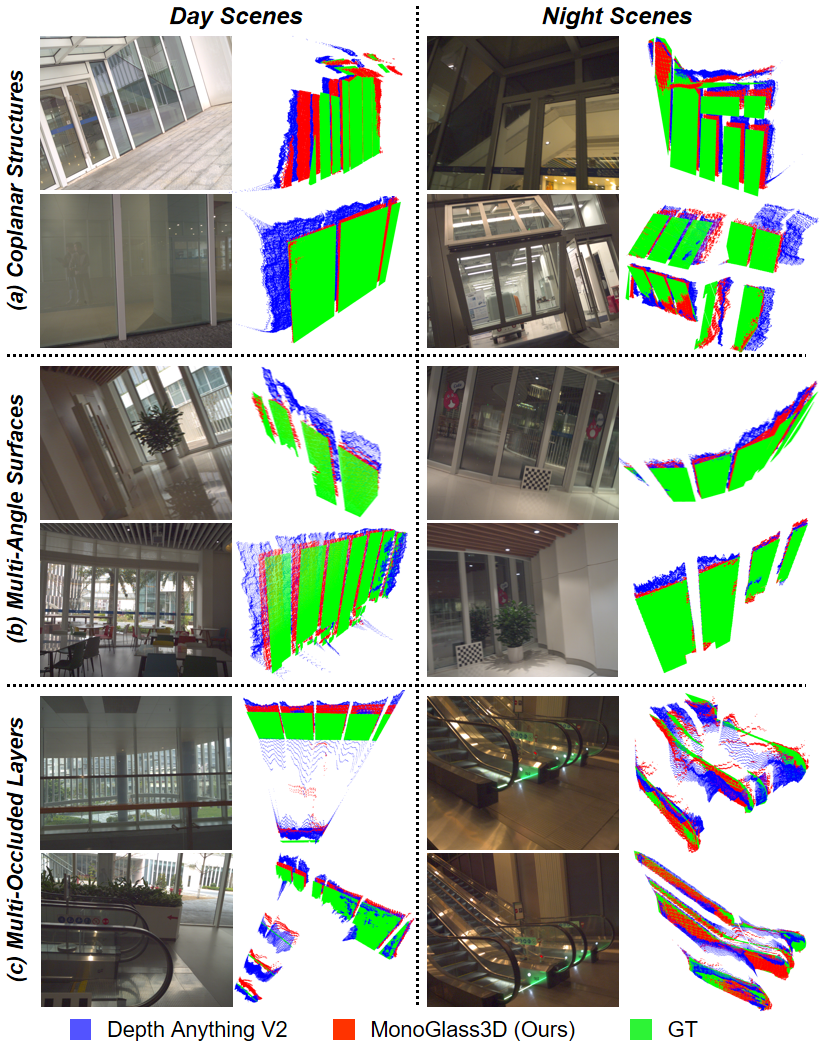}
    \caption{\textbf{Comparisons of glass detection results.} (a) Coplanar structures contain glass surfaces lying on same plane. (b) Multi-angle surfaces contain scenes with glass instances with relatively small angular difference. (c) Multi-occluded layers contain scenes with overlapping glass surfaces.}
    \label{fig:depth_samples}
    \vspace{-8pt}
\end{figure}

\begin{table*}[t!]
\vspace{-8pt}
\centering
\caption{Depth Estimation Metrics Comparison}
\renewcommand\arraystretch{1.3}
{
\begin{tabular}{ccccccccc}
\hline Test Data & Methods & Param. $\downarrow$ & Abs. Rel. $\downarrow$ & MAE $\downarrow$ & RMSE $\downarrow$  & $\sigma_1$ $\uparrow$ & $\sigma_2$ $\uparrow$ & $\sigma_3 $ $\uparrow$ \\
\hline
\multirow{4}{*}{All} & ZoeDepth (BEiT-B) \cite{zoedepth} & 112.0M & 0.156 & 0.720 & 0.806 & 0.757 & 0.937 & 0.989 \\
 & Depth Anything V2 (ViT-S) \cite{depth_anything_v2} & \textbf{24.8M} & 0.090 & 0.387 & 0.419 & 0.911 & 0.993 & 0.997 \\
 & Depth Anything V2 (ViT-B) \cite{depth_anything_v2} & 97.5M & 0.067 & 0.288 & 0.318 & 0.970 & 0.993 & 0.997\\
 & MonoGlass3D & 52.4M & \bf{0.063} & \bf{0.274} & \bf{0.298} & \bf{0.982} & \bf{0.994} & \bf{0.997} \\
\hline
\end{tabular}
\label{table:depth_eval}
}
\vspace{-10pt}
\end{table*}

\begin{figure*}[t]
    \centering
    \includegraphics[width=0.9\linewidth]{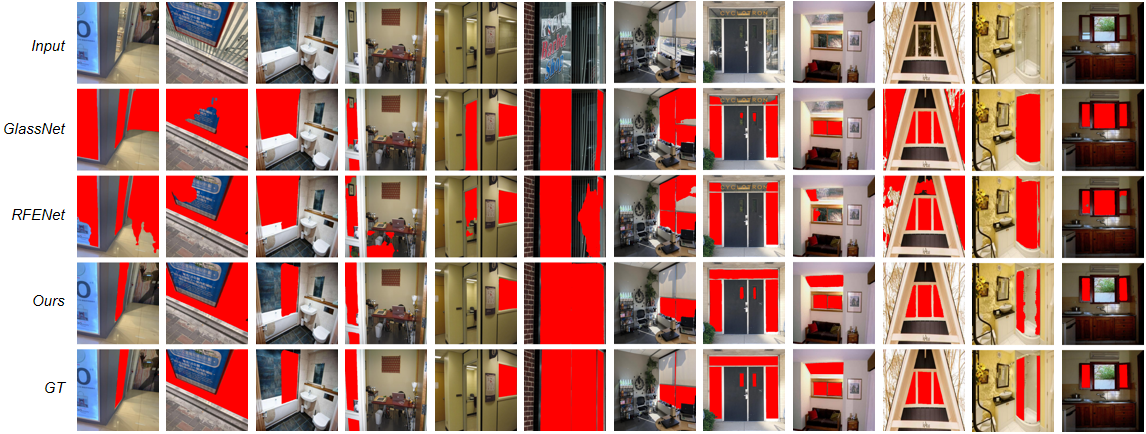}
    \caption{\textbf{Comparisons of glass segmentation results.} Compared to previous approaches, our method is better at accurately detecting glass in images, even when the surrounding context is ambiguous.}
    \label{fig:seg_res}
    \vspace{-15pt}
\end{figure*}

\subsubsection{Depth Estimation} 

For the depth estimation task, we finetuned ZoeDepth \cite{zoedepth} and Depth Anything V2 Metric \cite{depth_anything_v2} on our dataset. Quantitative results across various metrics are summarized in Table~\ref{table:depth_eval}, our method achieves the best performance under all evaluation metrics, and maintains stable performance in both day and night time data, despite having only $52.4\,\mathrm{M}$ trainable parameters, which is only $54\%$ of Depth Anything V2 (ViT-B). To facilitate a more intuitive comparison, we also visualize 3D projections of the estimated plane parameters, as shown in Fig.~\ref{fig:depth_samples}. For analysis, we categorize our dataset into three representative scene types.

\textit{Coplanar Structures} (Fig.~\ref{fig:depth_samples}(a)): This subset consists of images containing glass surfaces with regular geometric structures, often featuring multiple instances sharing identical plane parameters. In such cases, our approach demonstrates robust performance owing to the explicit plane parameter estimation, which ensures consistent and accurate surface fitting. Compared to direct depth regression, the depth maps projected from plane parameters are notably flatter and better aligned with the ground truth surfaces.

\textit{Multi-Angle Surfaces} (Fig.~\ref{fig:depth_samples}(b)): These examples contain multiple glass instances with minor angular differences between planes. Our method maintains stable and accurate fitting across surfaces with subtle variations, a task that is challenging for conventional depth estimation approaches \cite{zoedepth, depth_anything_v2} that lack explicit geometric supervision (e.g., normal vectors or plane parameters).

\textit{Multi-Occluded Layers} (Fig.~\ref{fig:depth_samples}(c)): This subset includes scenes with multiple overlapping glass surfaces, presenting the most challenging cases in our dataset. Our approach outperforms Depth Anything V2 in handling significant depth discontinuities. In the plane parameter space, parallel planes share the same normal but differ in intercepts, making such features more distinguishable and learnable by the network.

\begin{table}[h!]
\vspace{-10pt}
\centering
\caption{Glass Segmentation Comparison Results on the GDD Dataset}
\renewcommand\arraystretch{1.3}
{
\begin{tabular}{cccccc}
\hline Methods & Param. $\downarrow$ & IoU $\uparrow$ & $\mathrm{F1}\uparrow$ & MAE $\downarrow$ & BER $\downarrow$ \\
\hline GDNet \cite{gdnet} & 183.2M & 0.876 & 0.937 & 0.063 & 5.62 \\
 GlassNet \cite{gsd} & 83.7M & 0.881 & 0.932 & 0.059 & 5.71 \\
 EBLNet \cite{eblnet} & 46.2M & 0.887 & 0.940 & 0.055 & 5.36 \\
 GlassSemNet \cite{gsds} & 240.2M & 0.908 & 0.950 & 0.045 & 4.34 \\
 VBNet \cite{vbnet2024} & - & 0.907 & 0.948 & 0.048 & 4.70 \\
 MonoGlass3D & \bf{34.0M} & \bf{0.920} & \bf{0.951} & \bf{0.038} & \bf{3.76} \\
\hline
\end{tabular}
}
\label{table:gdd}
\vspace{-8pt}
\end{table}

\begin{table}[h!]
\centering
\caption{Glass Segmentation Comparisons Results on The GSD Dataset}
\renewcommand\arraystretch{1.3}
{\begin{tabular}{cccccc}
\hline Methods & Param. $\downarrow$ & IoU $\uparrow$ & $\mathrm{F1} \uparrow$ & MAE $\downarrow$ & BER $\downarrow$ \\
\hline GDNet \cite{gdnet} & 183.2M & 0.790 & 0.869 & 0.069 & 7.72 \\
GlassNet \cite{gsd} & 83.7M & 0.836 & 0.903 & 0.055 & 6.12 \\
GlassSemNet \cite{gsds} & 240.2M & 0.856 & 0.920 & 0.044 & 5.60 \\
RFENet \cite{rfenet} & 152.6M & 0.865 & \bf{0.931} & 0.048 & 6.23 \\
VBNet \cite{vbnet2024} & - & 0.861 & 0.921 & 0.043 & 5.51 \\
MonoGlass3D & \bf{34.0M} & \bf{0.872} & 0.917 & \bf{0.040} & \bf{5.16} \\
\hline
\end{tabular}
}
\label{table:gsd}
\vspace{-8pt}
\end{table}

\subsubsection{Glass Segmentation}
For a fair comparison with other methods, we disable the layers required for plane regression in our model, then trained and evaluated on two commonly used glass segmentation datasets: GDD \cite{gdnet} and GSD \cite{gsd}. Quantitative results are reported in Table~\ref{table:gdd} and Table~\ref{table:gsd}. Our model achieves state-of-the-art performance on both datasets, while using significantly fewer parameters compared to competing methods. Qualitative examples are shown in Fig.~\ref{fig:seg_res}. Compared to previous works that focus on specialized contextual cues, such as boundaries~\cite{eblnet, rfenet} or optical features~\cite{gwdepth, vbnet2024}, our approach leverages centerness maps for richer instance information and improved geometric context, enabling more accurate glass region identification, especially when boundaries are unclear or optical cues are weak.

\vspace{-5pt}
\subsection{Ablation Studies}

\begin{table*}[t]
\centering
{
\begin{threeparttable}[b]
\caption{Ablation Studies on Different Components}
\renewcommand\arraystretch{1.3}

\begin{tabular}{cccccccccccccccc}
\hline Base & AF & SA & CA & $L_{depth}$ & $L_{dist}$ & Cascades & Param. $\downarrow$ & IoU $\uparrow$ & Abs. Rel. $\downarrow$ & MAE $\downarrow$ & RMSE $\downarrow$ & $\sigma_1$ $\uparrow$ & $\sigma_2$ $\uparrow$ & $\sigma_3$ $\uparrow$\\
\hline
\ding{51} & & & & & \ding{51} & \ding{51} & \bf{31.15M} & 0.925 & 0.082 & 0.340 & 0.374 & 0.950 & 0.987 & 0.995 \\
\ding{51} & & & \ding{51} & & \ding{51} & \ding{51} & 38.97M & 0.924 & 0.072 & 0.317 & 0.346 & 0.972 & 0.993 & 0.997 \\
\ding{51} & & \ding{51} & \ding{51} & & \ding{51} & \ding{51} & 48.56M & 0.923 & 0.071 & 0.306 & 0.340 & 0.977 & 0.993 & 0.997 \\
\ding{51} & \ding{51} & \ding{51} & \ding{51} & \ding{51} & & \ding{51} & 52.43M & 0.940 & 0.073 & 0.313 & 0.336 & 0.975 & 0.993 & 0.997 \\
\ding{51} & \ding{51} & \ding{51} & \ding{51} & & \ding{51} & & 45.87M & 0.939 & 0.064 & 0.278 & 0.302 & 0.978 & 0.994 & 0.997 \\
\ding{51} & \ding{51} & \ding{51} & \ding{51} & & \ding{51} & \ding{51} & 52.43M & \bf{0.942} & \bf{0.063} & \bf{0.274} & \bf{0.298} & \bf{0.982} & \bf{0.994} & \bf{0.997} \\

\hline
\end{tabular}
\begin{tablenotes}
\item `Base' represents backbone and prediction heads, `AF' represents adaptive fusion modules, `SA' represents self-attention layers, `CA' represents cross-attention layers.
\end{tablenotes}
\label{table:abl_modules}
\end{threeparttable}
}
\vspace{-10pt}
\end{table*}

To evaluate the effectiveness of individual components in our proposed model, we conducted a series of ablation studies. The comparative results for both glass segmentation and depth estimation are presented in Table~\ref{table:abl_modules}. We begin with a `Base' model comprising only the backbone and prediction heads. Subsequently, we incrementally add the cross-attention (CA) module, self-attention (SA) module, and the adaptive fusion (AF) module to assess their respective contributions. 

\begin{figure}
    \centering
    \includegraphics[width=1\linewidth]{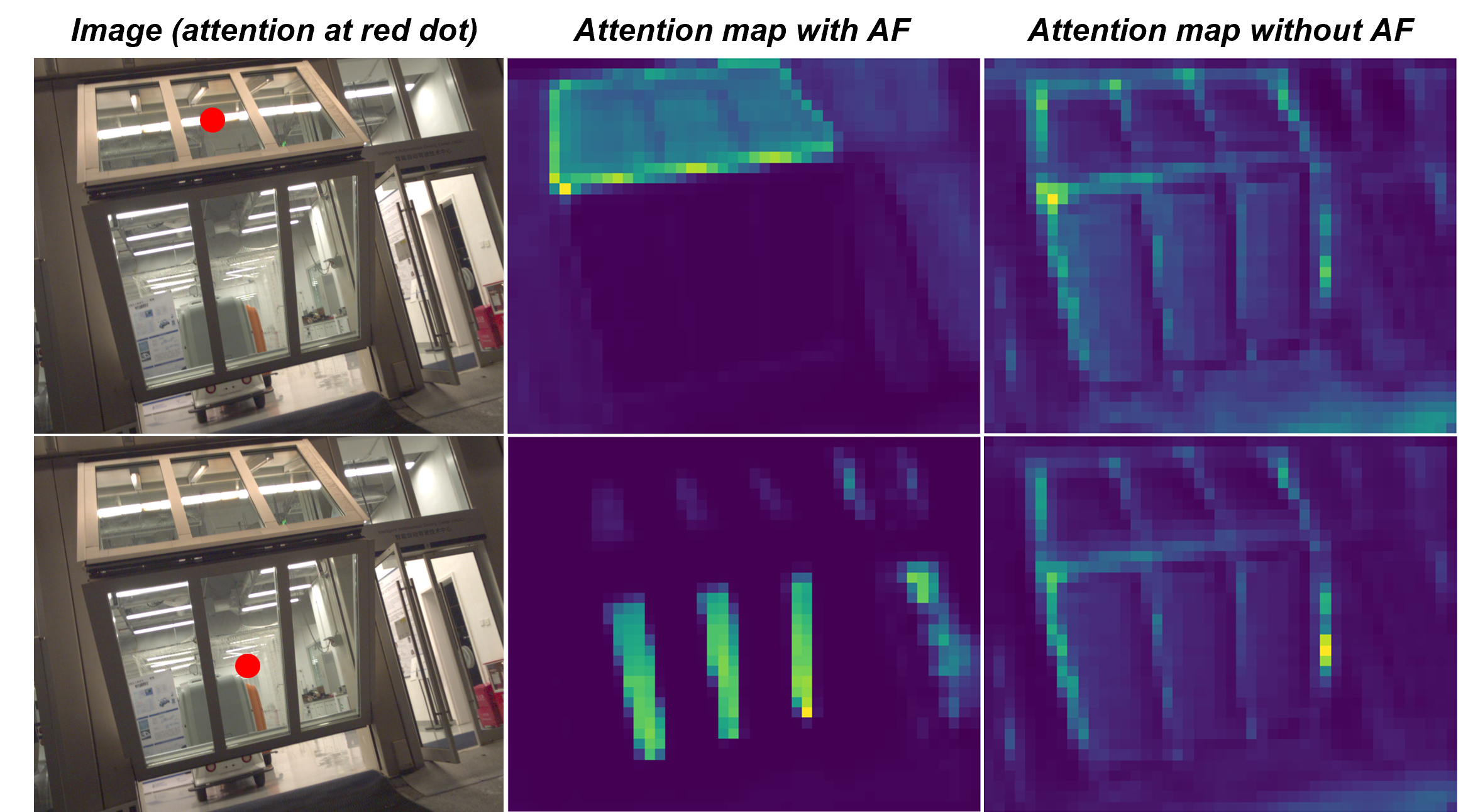}
    \caption{\textbf{Attention maps with and without AF.} The attention maps at the red dots show that AF modules enable the network to recognize different planes.}
    \label{fig:attmap}
    \vspace{-10pt}
\end{figure}

The results indicate that the transformer-based attention modules, CA and SA have limited impact on segmentation performance in IoU, but yield noticeable improvements in plane regression accuracy. The cascade structure is implemented to improve the learning efficiency of backbone layers, it also demonstrated noticeable improvements in overall performance. The AF module contributes the most significant performance gains among all components, which demonstrates the effectiveness of our centerness based adaptive feature fusion. To further illustrate its impact, attention maps generated with and without the AF module are compared in Fig.~\ref{fig:attmap}. The model enhanced with the AF module is able to more effectively capture the coplanar relations of the three glass windows at the top and the three at the bottom of the image.

Conventionally, segmentation and depth supervision do not provide explicit instance-level guidance, which is generally acceptable when object distinction is not required. In contrast, our approach seeks to distinguish glass surfaces residing on different planes, necessitating instance-level information. Centerness maps supply this instance-level knowledge without the computational overhead of explicit instance-level inference, which would unnecessarily complicate the problem.

We also evaluate the impact of different plane regression supervision strategies. As discussed in Section \ref{subsec:loss}, we argue that directly supervising plane parameter estimation using depth error is problematic due to its non-uniformity with respect to plane orientation. To address this, we adopt a loss based on the geometric distance between predicted and ground-truth planes. After retrained our network with depth loss $L_{depth}$, in place of plane distance loss $L_{dist}$, as shown in Table~\ref{table:abl_modules}, the results are consistent with our earlier analysis. The segmentation performance remains comparable, but depth estimation metrics deteriorate significantly. This confirms that the network is unable to achieve similar performance levels with depth supervision alone, further validating the effectiveness of our proposed plane distance loss.

\subsection{Real Scene Experiment}

\begin{figure}
    \centering
    \includegraphics[width=1\linewidth]{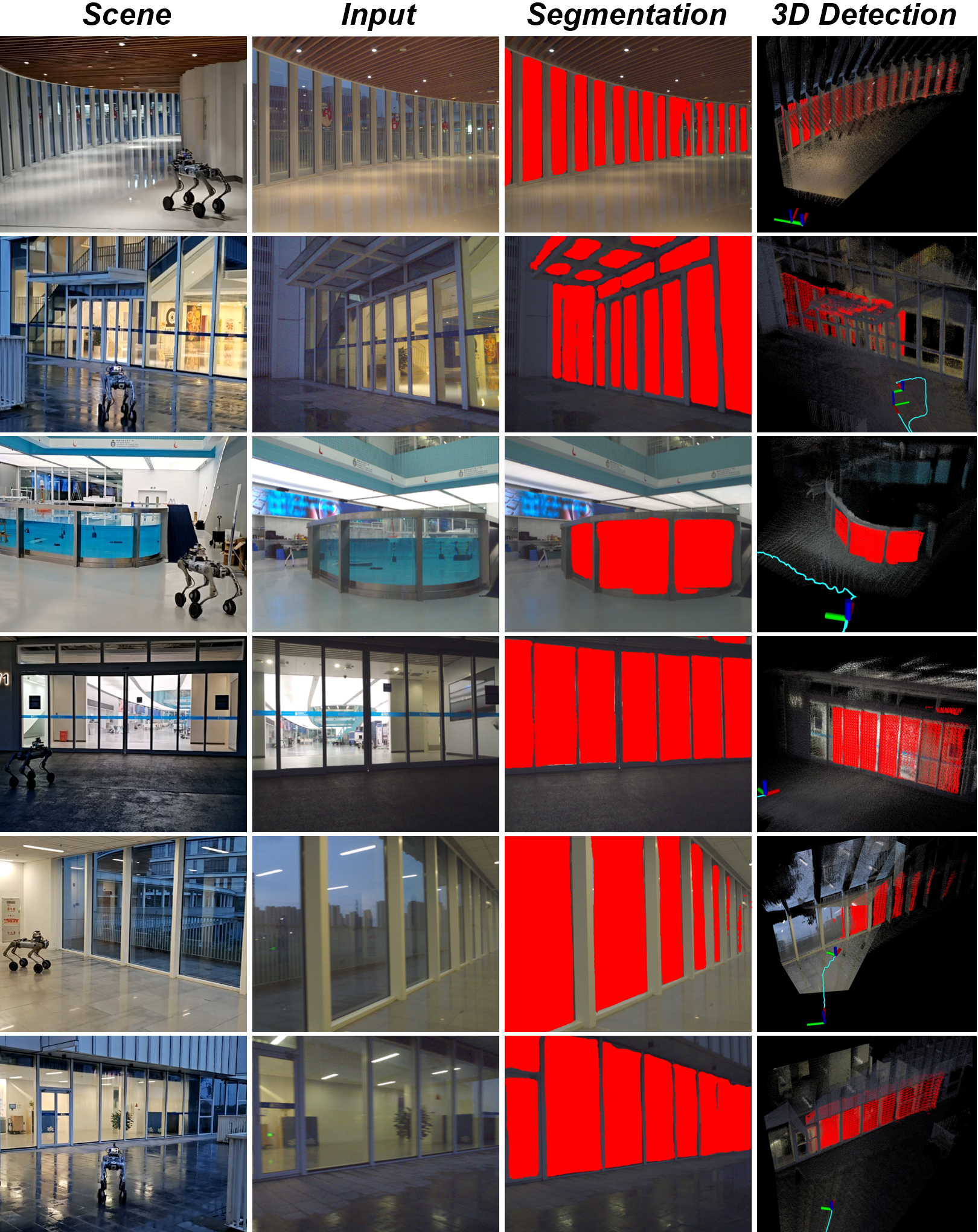}
    \caption{\textbf{3D glass detection results in real scenes.} We evaluate our approach in scenes that are absent in the training set, and it maintains strong performance.}
    \label{fig:real_tests}
    \vspace{-10pt}
\end{figure}

The ultimate goal of this research is to provide a practical monocular 3D glass detection solution. To further assess real-world performance, from our dataset, we deliberately select six scenes in which all frames are excluded from the training set, including indoor and outdoor scenes with different illumination conditions, resulting in a total of 180 images. This experimental setup allows us to rigorously evaluate the generalization capability of our model by explicitly testing it on entirely unseen environments. The results on depth metrics are presented in Table~\ref{table:depth_unseen}, with segmentation performance of $0.940$ IoU. Some result samples are shown in Fig.~\ref{fig:real_tests}. Although our model's performance decreases compared to evaluations where most scenes are included in training, it still consistently outperforms competing methods on these challenging samples. Our approach demonstrated stable and reliable performance, particularly when detecting large glass surfaces at close range ($<10$ meters). However, false negatives are more likely to occur for glass surfaces located farther from the camera. We attribute this limitation to the predominance of short-range data in our training set, which constrains the model’s ability to effectively detect distant glass surfaces.

\begin{table}
\centering
\caption{Depth Estimation Metrics Comparison on Real Scenarios}
\renewcommand\arraystretch{1.5}
{
\begin{tabular}{ccccc}
\hline 
Methods & Abs. Rel. $\downarrow$ & MAE $\downarrow$ & RMSE $\downarrow$  & $\sigma_1$ $\uparrow$ \\
\hline
ZoeDepth (BEiT-B) \cite{zoedepth} & 0.215 & 1.056 & 1.159 & 0.600 \\
Dep. Any. V2 (ViT-S) \cite{depth_anything_v2} & 0.123 & 0.547 & 0.577 & 0.841 \\
Dep. Any. V2 (ViT-B) \cite{depth_anything_v2} & 0.088 & 0.386 & 0.409 & 0.952 \\
MonoGlass3D & \bf{0.083} & \bf{0.383} & \bf{0.405} & \bf{0.976} \\
\hline
\end{tabular}
\label{table:depth_unseen}
}
\vspace{-8pt}
\end{table}

\subsection{Discussion}

\subsubsection{Limitations}

Our method works on the basis of assuming all glass surfaces can be approximated as 3D planes, which does not always hold in real-world scenarios. In practice, glass surface shapes follow a long-tailed distribution: while the majority can be reasonably approximated as planes, there remain numerous cases with curved or irregular geometries. This intrinsic variability presents a significant challenge for designing a 3D glass detection method that generalizes to arbitrary glass shapes.

Additionally, despite our efforts to streamline data collection and annotation, constructing our dataset remains a labor-intensive process. This raises challenges for scaling up the dataset to cover a wider variety of glass scenarios. The primary difficulty arises from the need to annotate glass surfaces both in 3D point clouds and in 2D RGB images, and then accurately associate these labels across modalities. Since the 3D positions of glass are not directly observable, we have to rely on supplementary data modalities and manual annotation to achieve precise 3D labeling.

\subsubsection{Future Work}

While end-to-end monocular 3D glass detection is conceptually simple and elegant, monocular RGB cameras inherently suffer from limited 3D perception capabilities, primarily due to scale ambiguity and susceptibility to varying lighting conditions. For increased robustness and accuracy in 3D perception, it is desirable to incorporate data from additional sensor modalities. In future work, we plan to investigate the integration of multi-modal inputs—such as depth sensors, LiDAR, or stereo vision—to further enhance the performance and reliability of 3D glass detection systems.

\section{Conclusion}

In this work, we address the challenging problem of 3D glass detection with monocular input. To facilitate this research, we build a novel dataset featuring complex glass structures with precise annotations. We propose MonoGlass3D, a unified network performs glass segmentation and plane regression simultaneously. Central to our design is the Adaptive Fusion module, which leverages centerness maps to provide geometric context guidance, significantly enhancing both segmentation and plane parameter estimation. We formulate 3D glass detection as a plane regression problem, which allows us to exploit the inherent geometric properties of 3D planes for more effective learning. Extensive experiments demonstrate that our approach achieves state-of-the-art performance in both glass segmentation and glass depth estimation.

\bibliographystyle{IEEEtran}
\bibliography{references}
\vspace{-10pt}
\begin{IEEEbiography}[{\includegraphics[width=1in,height=1.25in,clip,keepaspectratio]{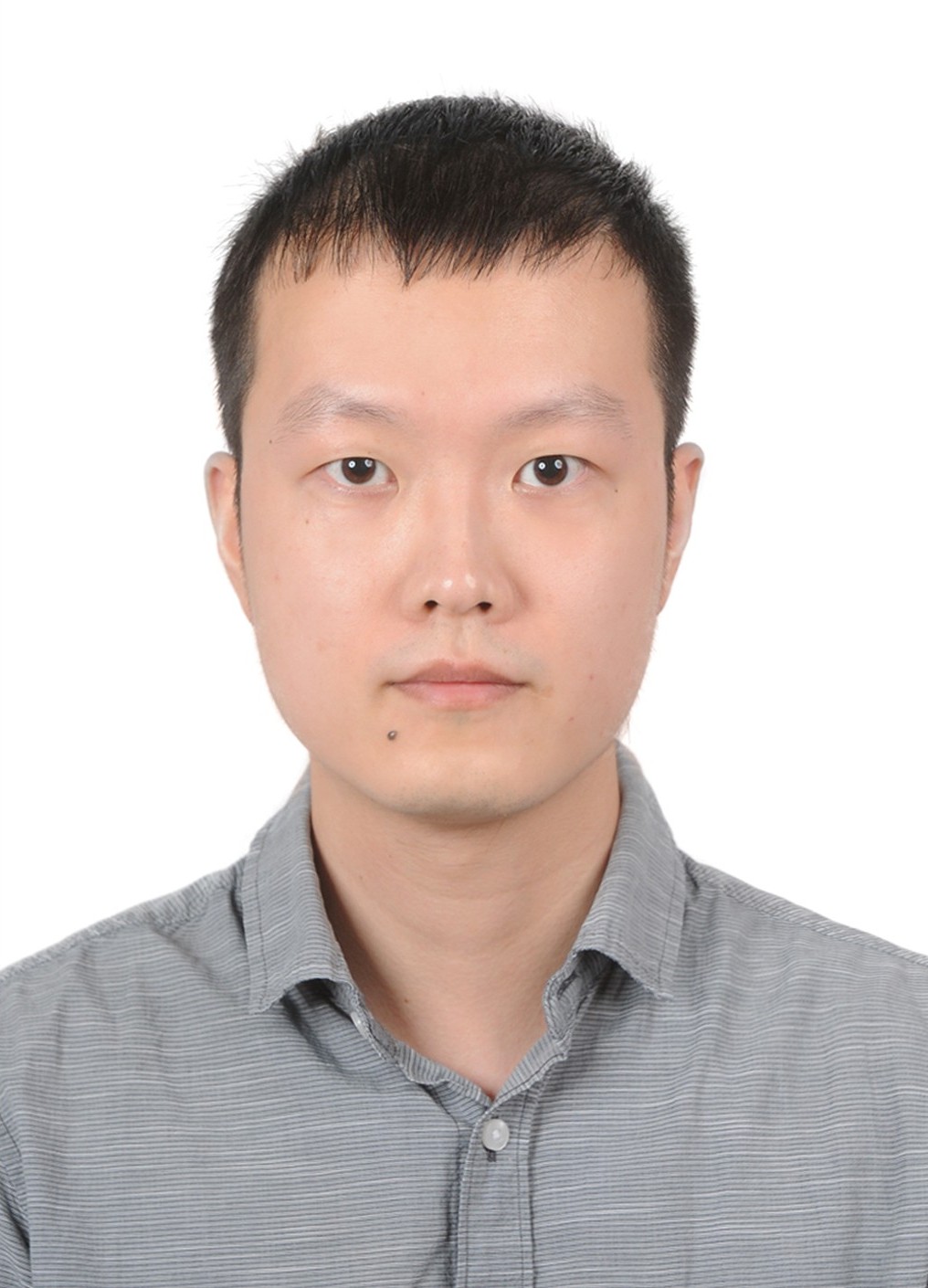}}]{Kai Zhang}
(Student Member, IEEE) Recieved the B.Eng. degree in mechatronics engineering from the University of Sydney, Sydney, Australia, in 2020 and the M.Phil. degree in robotics and autonomous systems in the The Hong Kong University of Science and Technology (Guangzhou), Guangzhou, China, in 2024, where he is currently working toward the Ph.D. degree in robotics and autonomous systems at the Intelligent Autonomous Driving Center. His research interests include robotics perception, SLAM, and embodied intelligence.
\end{IEEEbiography}

\begin{IEEEbiography}[{\includegraphics[width=1in,height=1.25in,clip,keepaspectratio]{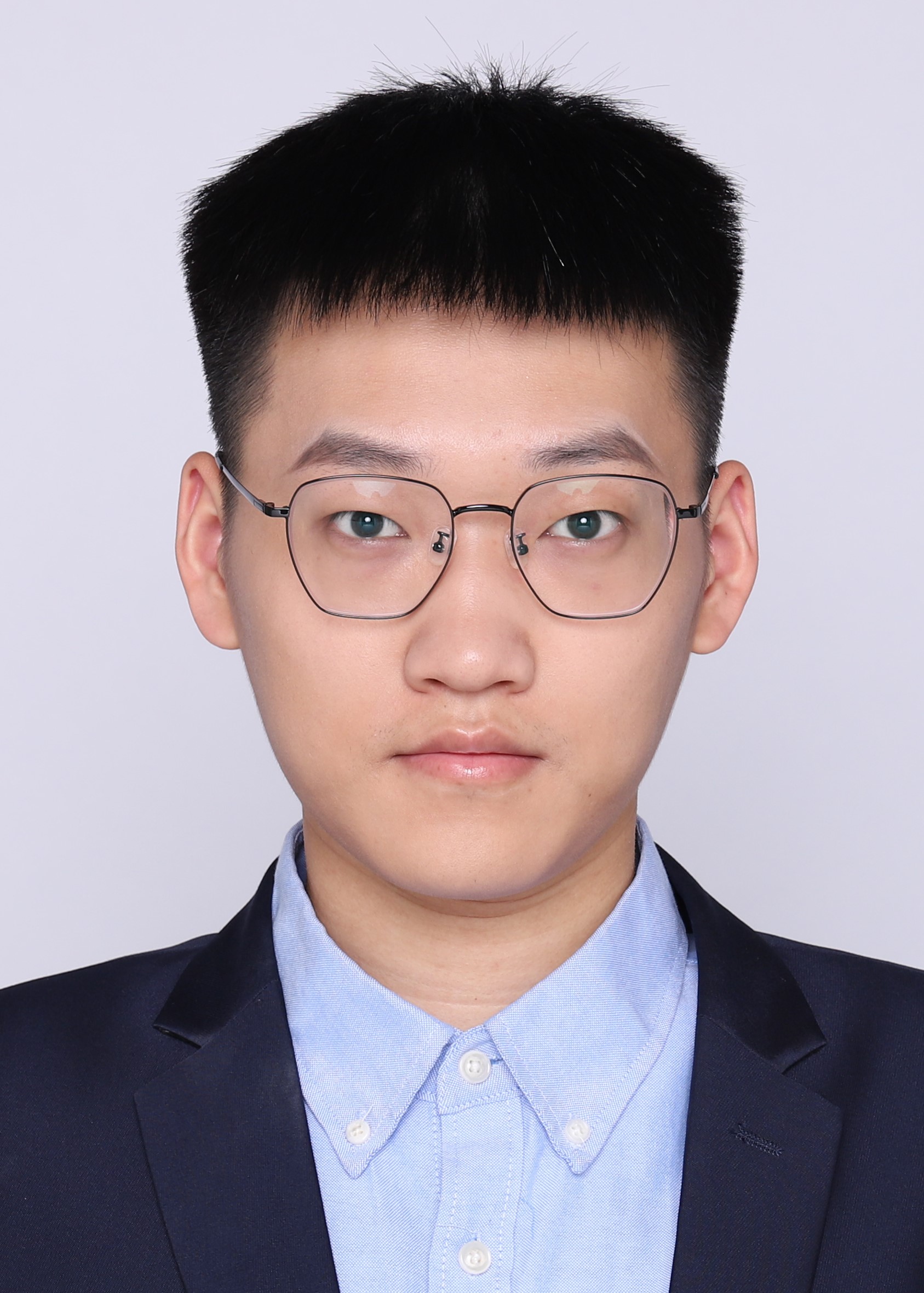}}]
{Guoyang Zhao} (Student Member, IEEE) received the B.Eng. degree in logistics engineering from Northeast Agricultural University, Harbin, China, in 2022, and the M.Phil. degree in robotics and autonomous systems from The Hong Kong University of Science and Technology (Guangzhou), Guangzhou, China, in 2024. He is currently pursuing the Ph.D. degree at the Intelligent Autonomous Driving Center, Robotics and Autonomous Systems Thrust, The Hong Kong University of Science and Technology, Guangzhou, China. His research interests include computer vision, robotics navigation, and deep learning.
\vspace{-25pt}
\end{IEEEbiography}

\begin{IEEEbiography}[{\includegraphics[width=1in,height=1.25in,clip,keepaspectratio]{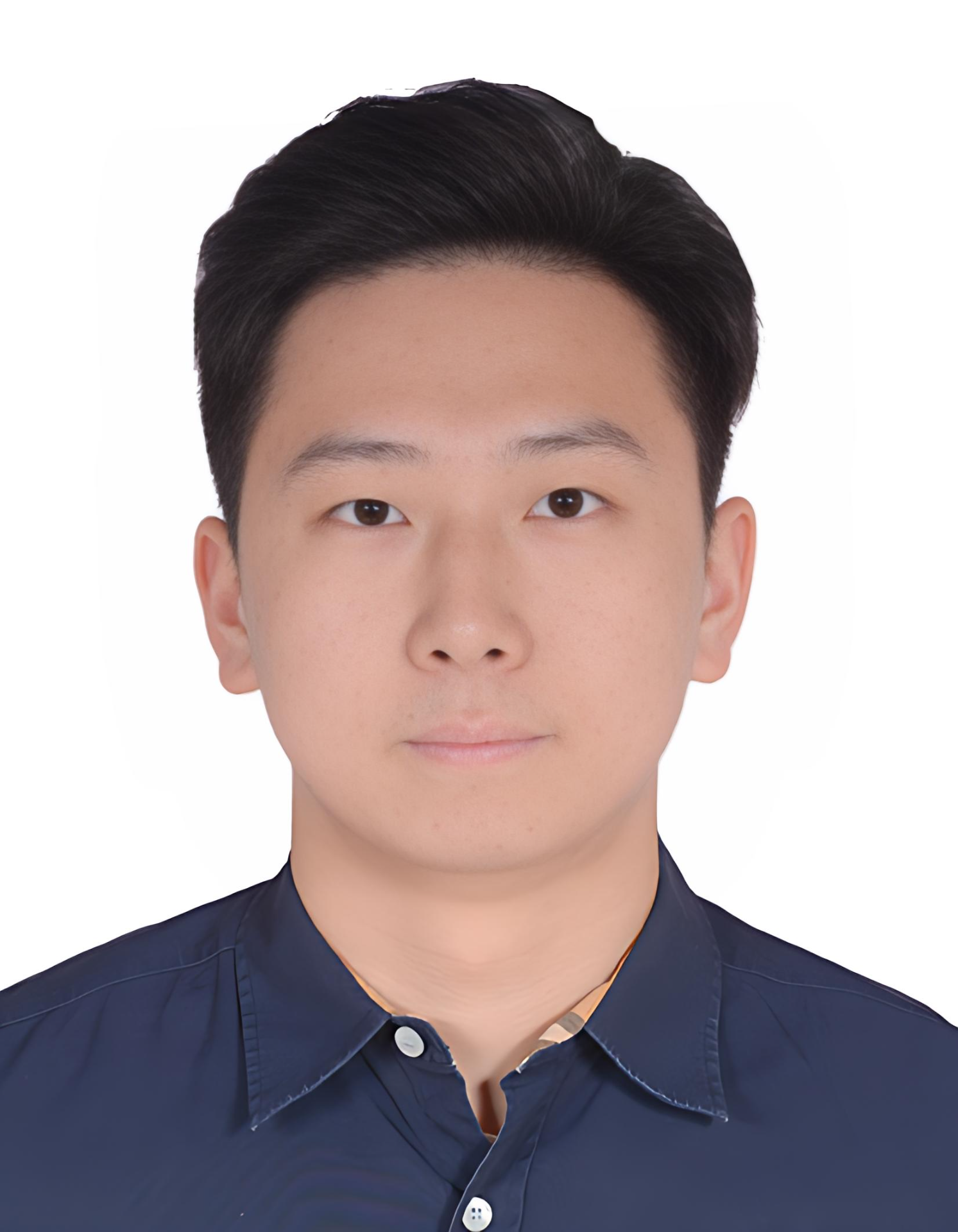}}]{Jianxing Shi}
(Student Member, IEEE) received his B.Eng. degree in Mechanical Engineering from Shandong University in 2022 and his M.Phil. degree in Robotics and Autonomous Systems from The Hong Kong University of Science and Technology (Guangzhou) in 2025. His research interests include sensor fusion, LiDAR SLAM, and robotic perception.
\vspace{-25pt}
\end{IEEEbiography}

\begin{IEEEbiography}[{\includegraphics[width=1in,height=1.25in,clip,keepaspectratio]{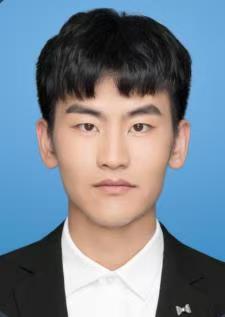}}]
{Bonan Liu} (Student Member, IEEE) received the bachelor's degree in engineering mechanic from Central South University ChangSha, China, in 2020, the master's degree in Data Science from the City university of Hong Kong in 2022. He is currently a Ph.D. candidate student in the information hub in HKUST-GZ.
\vspace{-25pt}
\end{IEEEbiography}

\begin{IEEEbiography}[{\includegraphics[width=1in,height=1.25in,clip,keepaspectratio]{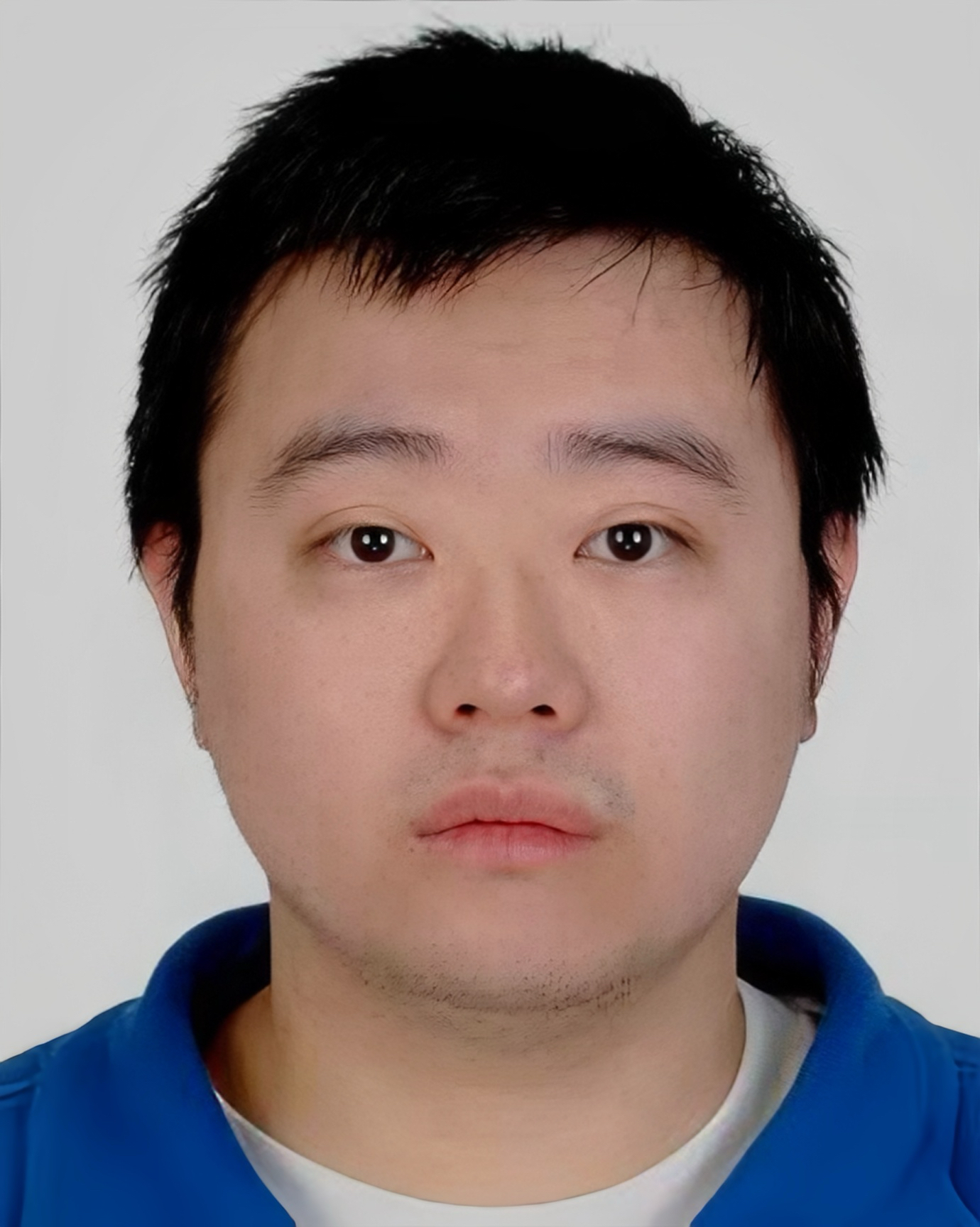}}]{Weiqing Qi} received the B.S. degree in Computer Science from University of California, Santa Barbara, CA, USA, in 2021, and the M.Phil. degree in robotics and autonomous systems from The Hong Kong University of Science and Technology (Guangzhou), Guangzhou, China, in 2024. His current research interests include lane detection, drivable area segmentation, and semantics segmentation, etc.
\vspace{-25pt}
\end{IEEEbiography}

\begin{IEEEbiography}[{\includegraphics[width=1in,height=1.25in,clip,keepaspectratio]{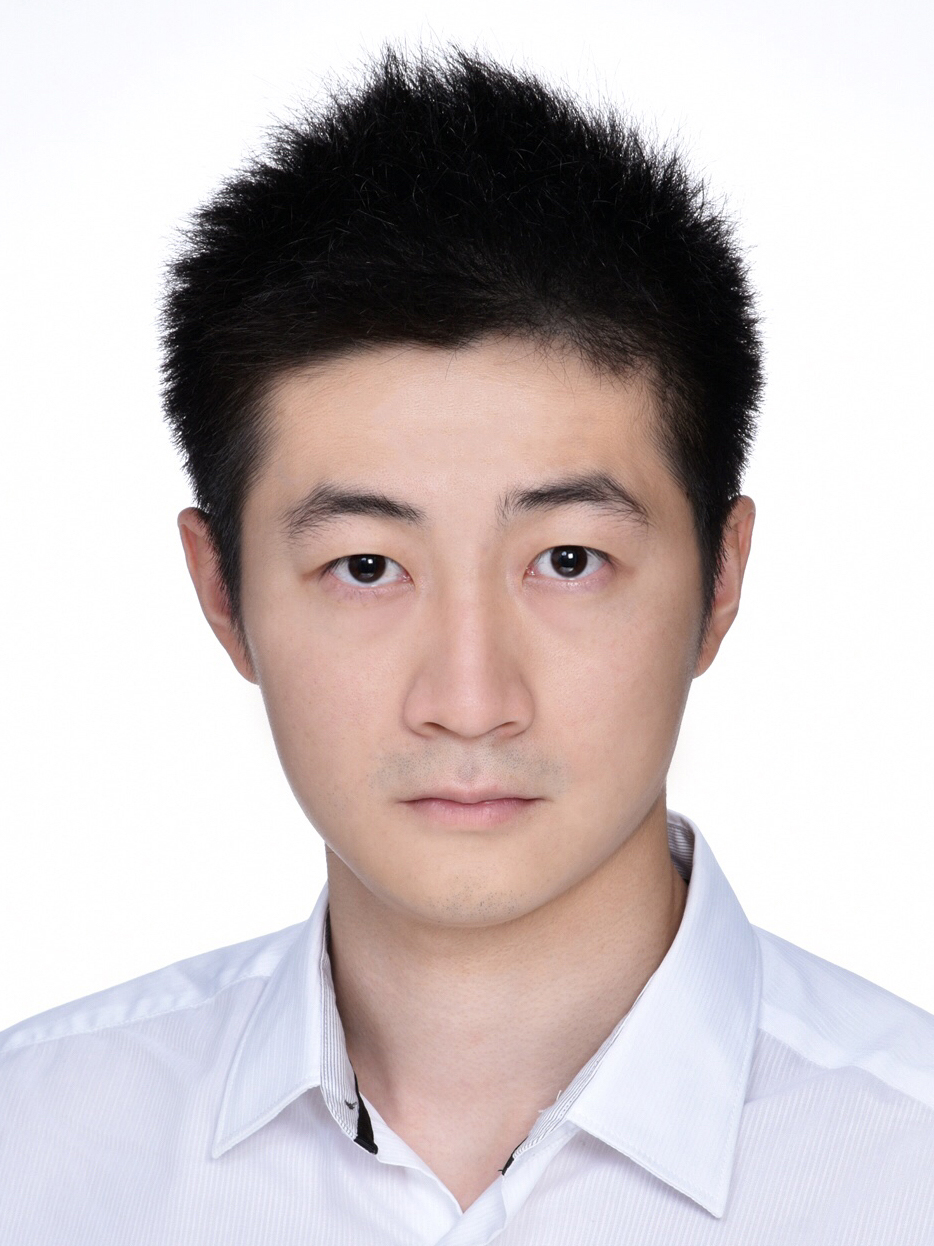}}]
{Jun Ma} (Senior Member, IEEE) received the B.Eng. degree with First Class Honours in electrical and electronic engineering from  Nanyang Technological University, Singapore, in 2014, and the Ph.D. degree in electrical and computer engineering from the National University of Singapore, Singapore, in 2018.
From 2018 to 2021, he held several positions at the National University of Singapore; University College London, London, U.K.; University of California, Berkeley, Berkeley, CA, USA; and Harvard University, Cambridge, MA, USA.  He is currently an Assistant Professor with the Robotics and Autonomous Systems Thrust, The Hong Kong University of Science and Technology (Guangzhou), Guangzhou, China, and also with the Division of Emerging Interdisciplinary Areas, The Hong Kong University of Science and Technology, Hong Kong SAR, China. He is also the Director of Intelligent Autonomous Driving Center, The Hong Kong University of Science and Technology (Guangzhou), Guangzhou, China.
His research interests include motion planning and control for robotics and autonomous driving.
\end{IEEEbiography}

\end{document}